\documentclass[letterpaper]{article} % DO NOT CHANGE THIS
\usepackage{aaai2026}  % DO NOT CHANGE THIS
\usepackage{times}  % DO NOT CHANGE THIS
\usepackage{helvet}  % DO NOT CHANGE THIS
\usepackage{courier}  % DO NOT CHANGE THIS
\usepackage[hyphens]{url}  % DO NOT CHANGE THIS
\usepackage{graphicx} % DO NOT CHANGE THIS
\usepackage{natbib}  % DO NOT CHANGE THIS AND DO NOT ADD ANY OPTIONS TO IT
\usepackage{caption} % DO NOT CHANGE THIS AND DO NOT ADD ANY OPTIONS TO IT
\usepackage{algorithm}
\usepackage{amssymb}
\usepackage{amsmath}
\usepackage{booktabs}
\usepackage{array}
\usepackage{url}
\usepackage{array}
\usepackage{booktabs}
\usepackage{arydshln}
\usepackage{multirow}
\usepackage{ragged2e}
\usepackage{amsmath}       
\usepackage{algpseudocode} 

\usepackage{newfloat}
\usepackage{listings}
\DeclareCaptionStyle{ruled}{labelfont=normalfont,labelsep=colon,strut=off} % DO NOT CHANGE THIS
\floatstyle{ruled}
\newfloat{listing}{tb}{lst}{}
\floatname{listing}{Listing}
\title{Induce, Align, Predict: Zero-Shot Stance Detection via Cognitive Inductive Reasoning}

\author{
    Bowen Zhang\textsuperscript{\rm 1},
    Jun Ma\textsuperscript{\rm 1},
    Fuqiang Niu\textsuperscript{\rm 2},
    Li Dong\textsuperscript{\rm 1},
    Jinzhou Cao\textsuperscript{\rm 1},
    Genan Dai\textsuperscript{\rm 1}\thanks{Corresponding author.}
}
\affiliations{
    \textsuperscript{\rm 1}School of Artificial Intelligence, Shenzhen Technology University\\
    \textsuperscript{\rm 2}School of Cyber Science and Technology, University of Science and Technology of China\\
    zhang\_bo\_wen@foxmail.com, 2300411011@email.szu.edu.cn, nfq729@gmail.com, dongli@sztu.edu.cn,
    caojinzhou@sztu.edu.cn,
    daigenan@sztu.edu.cn
}

\usepackage{bibentry}
\begin{document}

\maketitle

\begin{abstract}
Zero-shot stance detection (ZSSD) seeks to determine the stance of text toward previously unseen targets, a task critical for analyzing dynamic and polarized online discourse with limited labeled data. While large language models (LLMs) offer zero-shot capabilities, prompting-based approaches often fall short in handling complex reasoning and lack robust generalization to novel targets. Meanwhile, LLM-enhanced methods still require substantial labeled data and struggle to move beyond instance-level patterns, limiting their interpretability and adaptability. 
Inspired by cognitive science, we propose the Cognitive Inductive Reasoning Framework (CIRF), a schema-driven method that bridges linguistic inputs and abstract reasoning via automatic induction and application of cognitive reasoning schemas. CIRF abstracts first-order logic patterns from raw text into multi-relational schema graphs in an unsupervised manner, and leverages a schema-enhanced graph kernel model to align input structures with schema templates for robust, interpretable zero-shot inference. 
Extensive experiments on SemEval-2016, VAST, and COVID-19-Stance benchmarks demonstrate that CIRF not only establishes new state-of-the-art results, but also achieves comparable performance with just 30\% of the labeled data, demonstrating its strong generalization and efficiency in low-resource settings. 
\end{abstract}


\section{Introduction}
 
Zero-shot stance detection (ZSSD) aims to identify the stance of text towards targets unseen during training, a challenge increasingly important for analyzing polarized social media discourse as new topics rapidly emerge and labeled data remain scarce~\cite{allaway2020zero, liang2022zero}.
Recent advances in large language models (LLMs) offer new opportunities for ZSSD, as their zero-shot prompting and strong contextual understanding enable deeper semantic reasoning and generalization to novel targets~\cite{binz2023using}.
However, LLMs applied via prompting strategies~\cite{li2023stance,zhang2023investigating,lan2024stance,zhao2024zerostance,weinzierl2024tree}
typically exhibit suboptimal performance on ZSSD. In contrast, LLM-enhanced methods (LEM) combine the rich knowledge encoded in LLMs with the task-specific adaptation capabilities of supervised neural models, and have recently become a prominent research direction
~\cite{zhang2024knowledge,dai2025large,zhang2025logic}.
Nevertheless, these approaches still face two challenges: substantial reliance on labeled data and limited generalization beyond surface-level lexical cues~\cite{wang2020neural}.

On the other hand, cognitive science suggests that humans reason not only by memorizing specific examples, but also by abstracting generalizable reasoning schemas—structured templates that capture the logical relationships underlying diverse arguments (e.g., causality, value judgments, conditional inference) \cite{halpern1990analysis,barwise1977introduction,tenenbaum2011grow}. For instance, arguments such as “increases health risks” and “reduces economic stability” both instantiate a schema of negative consequence, leading to \textit{opposition}, regardless of surface wording or topical context. Such schemas disentangle reasoning logic from lexical or contextual specifics, enabling robust generalization to novel targets \cite{cocchi2013dynamic,ragni2013theory}. This cognitive insight motivates us to integrate schema-level reasoning into stance detection, aiming to overcome the data dependency and limited generalization of current instance-based approaches.

% Reasoning schemas have demonstrated strong potential for enhancing both generalization and interpretability across various domains, such as question answering, causal inference, and narrative understanding~\cite{peng2024hypothesis, li2024schema, cheng2024shield, su2025enhancing,tao2025comprehensive}. 
% However, the application of schema-based reasoning to stance detection, especially in the zero-shot setting, remains largely unexplored. 
% This gap is primarily attributable to two key challenges: (1) the absence of effective methods for modeling and extracting stance-specific reasoning schemas that generalize across diverse targets, and (2) the lack of mechanisms to align input instances with abstract schemas, thereby supporting schema-guided inference and robust generalization to novel stance targets.
Reasoning schemas have shown strong potential for improving generalization and interpretability in domains such as question answering and causal inference~\cite{peng2024hypothesis, li2024schema, cheng2024shield, su2025enhancing,tao2025comprehensive}. However, their application to stance detection—especially in the zero-shot setting—remains largely unexplored. This is mainly due to two core challenges:
(1) the lack of effective methods for modeling and extracting stance-specific reasoning schemas that can generalize across diverse targets; and
(2) the absence of mechanisms for aligning input instances with abstract schemas to enable schema-guided inference.
Existing neural and LLM-based approaches predominantly rely on instance-level lexical or contextual patterns, rarely inducing or leveraging reusable schema-level reasoning. This limits both their generalization and interpretability in ZSSD.

To address these challenges, we propose the Cognitive Inductive Reasoning Framework (CIRF), a schema-driven approach for ZSSD that bridges linguistic input and abstract reasoning via automatic schema induction and schema-guided inference. Specifically, CIRF comprises two key components:
(1) Unsupervised Schema Induction (USI), which leverages LLMs to abstract structured reasoning patterns from raw text by converting predicates into first-order logic (FOL) expressions and clustering them into a multi-relational schema graph, capturing stance-relevant logical relations independent of specific targets or vocabulary; and
(2) Schema-Enhanced Graph Kernel Model (SEGKM), which represents each input as an FOL graph, maps predicate nodes to corresponding schema nodes, and employs a learnable graph kernel to align input structures with schema templates for stance prediction.
Unlike standard graph neural networks, which often struggle to capture reusable high-order reasoning motifs, our kernel-based approach enables direct and interpretable alignment between input reasoning structures and abstract schema templates. This explicit structural and semantic matching is essential for robust zero-shot generalization and interpretability in stance detection, and cannot be easily achieved by conventional neural architectures.
This framework enables CIRF to generalize effectively to novel and diverse targets by combining the interpretability of symbolic schemas with the adaptability of neural inference, setting a new paradigm for schema-driven zero-shot stance detection.

In summary, the contributions of this work are as follows:

\begin{itemize}

\item We propose the CIRF, a first schema-driven approach for ZSSD that bridges linguistic input and abstract reasoning via automatic induction and application of cognitive reasoning schemas.

\item  We introduce a unified framework combining unsupervised schema induction—leveraging LLMs to abstract FOL patterns into multi-relational schema graphs—and SEGKM for effective schema-guided stance inference and unseen target generalization.

\item  Extensive experiments on the SemEval‑2016, VAST, and COVID‑19‑Stance benchmarks demonstrate that CIRF outperforms state‑of‑the‑art ZSSD baselines and achieves competitive results with 70\% fewer labeled examples than LLM-enhanced methods.
% (3) Extensive experiments on the SemEval-2016, VAST and COVID-19-Stance benchmarks demonstrate the superiority of CIRF: it outperforms state-of-the-art ZSSD baselines by 1.9, 0.6 and 3.7 percentage points in macro-F1, respectively, and achieves competitive performance with 70\% fewer labeled examples compared to LLM-enhanced methods. 
\end{itemize}

\section{Related Work}

% \paragraph{ZSSD Methods.}
\textbf{ZSSD Methods.}
ZSSD has attracted increasing attention due to its importance in identifying stances toward previously unseen targets~\cite{liang2022zero}. 
Early approaches, such as JointCL~\cite{liang2022jointcl} and TarBK~\cite{zhu2022enhancing}, rely heavily on supervised learning with large annotated datasets, limiting their generalization to novel targets. 
Recent advances in LLMs have introduced new paradigms for ZSSD, including zero-shot prompting LLMs~\cite{zhang2023investigating,lan2024stance} and LLM-enhanced fine-tuned models~\cite{li2023stance,zhang2024knowledge,dai2025large,zhang2025logic}. 
However, zero-shot prompting LLMs often underperform due to their lack of task-specific adaptation, while LLM-enhanced fine-tuned models still require extensive instance-level supervision. These limitations highlight the need for frameworks that can generalize reasoning patterns to unseen targets without relying on large amounts of labeled data, thereby motivating schema-driven approaches.

% \paragraph{First-Order Logic for Neural Reasoning.}
\textbf{First-Order Logic for Neural Reasoning.}
FOL provides a structured and interpretable foundation for encoding logical relations such as causality, implication, and conditionality, and has been widely adopted to enhance consistency and transparency in neural reasoning~\cite{hu2016deep,huang2022logic}. Recent methods integrate FOL constraints into neural architectures via posterior regularization~\cite{hu2016harnessing,zhang2022sentiment} or joint FOL-neural embeddings, aiming to unify symbolic rigor with statistical flexibility. 
In stance detection, recent studies~\cite{dai2025large,zhang2025logic} prompt LLMs to generate FOL-based reasoning chains, achieving notable improvements over conventional models. However, such FOL-based techniques typically rely on instance-specific rules, limiting their ability to induce domain-agnostic abstractions or generalize logic across unseen topics—a critical bottleneck in zero-shot scenarios. 
Moreover, manually crafted or instance-specific FOL rules often struggle to capture high-level reasoning schemas that generalize across domains or topics, an aspect critical for ZSSD.

\textbf{Schema Induction Methods.}
Schema induction has been widely explored in event-centric scenarios~\cite{edwards-ji-2023-semi,huang2016liberal,shen2021corpus}, where both bottom-up concept linking and top-down clustering are employed to capture relational patterns. The advent of LLMs has further enabled schema construction through generative and summarization capabilities~\cite{li2023opendomain,tang2023harvesting,dror2023zero,shi2024language,rong2025pred,finch2025generative,lee2025few}. However, most existing approaches focus on multi-sentence or event-level analysis, making them ill-suited for sentence-level stance detection, which requires inferring implicit and diverse semantic relations from less structured input.
Among recent efforts, SenticNet8~\cite{cambria2024senticnet} and LogiMDF~\cite{zhang2025logic} are most relevant, aiming to abstract lexical items or induce logical rules via LLMs. Yet, these methods fundamentally operate at the word or predicate level and largely overlook the relational structures that govern inter-concept dependencies—such as causality or contradiction—that are vital for robust reasoning. As a result, they often yield semantically shallow or fragmented schema representations, limiting generalization and structural expressivity.
In contrast, our CIRF treats entire FOL reasoning chains as the unit of schema induction. This allows us to abstract and transfer not only individual semantic items but also the relational patterns binding them, enabling more expressive and transferable schemas for zero-shot stance detection.

\section{Method}
\begin{figure*}[t] % h 表示尝试将图片放在当前位置，顶，底，单独页面
    \centering % 居中显示图片
    \includegraphics[width=0.8\textwidth]{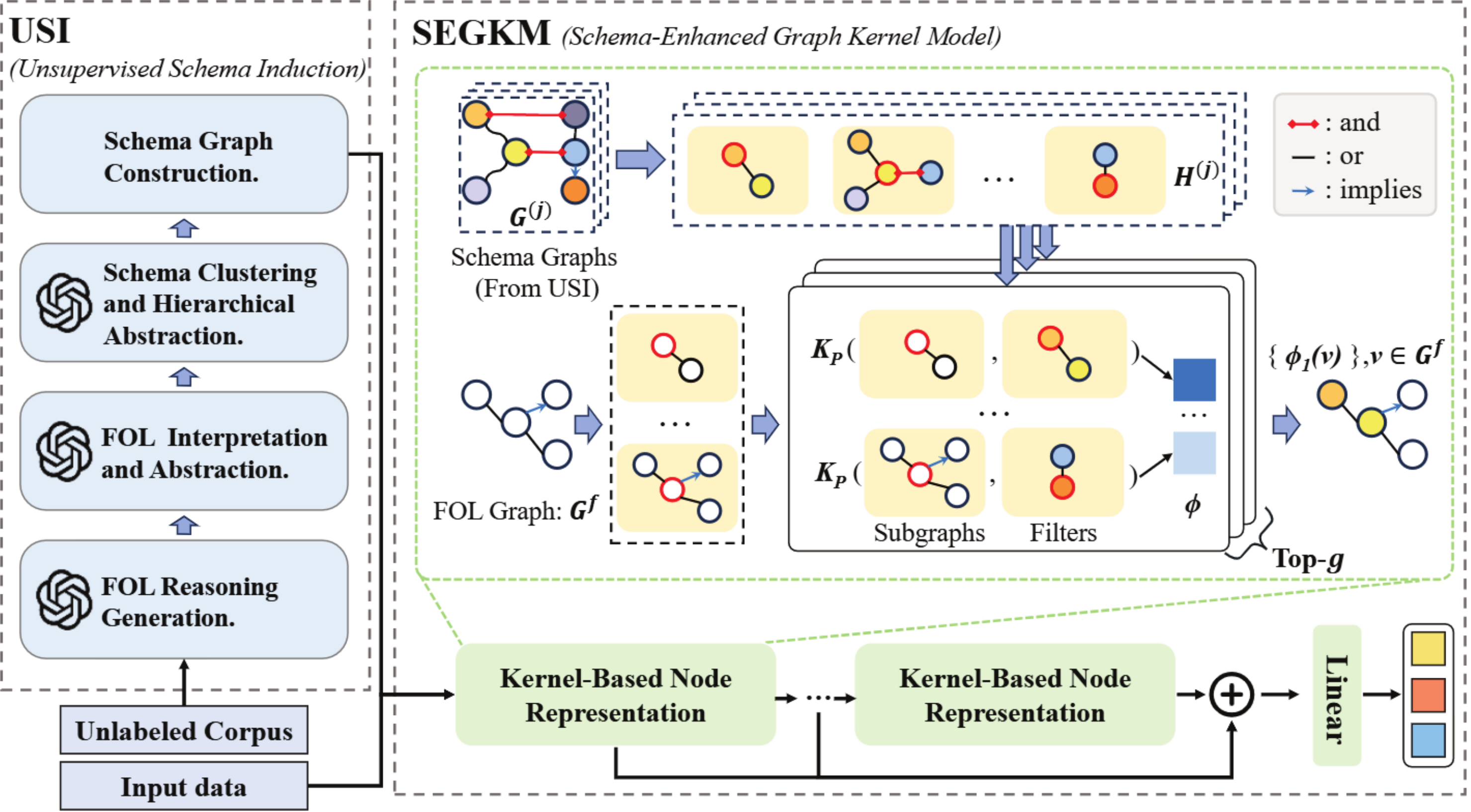}
    \caption{The framework of CIRF.
The red-bordered circles represent the central nodes of subgraphs or filters.
    }
    \label{fig:framework} % 设置图片标签，用于引用
\end{figure*}

Our CIRF tackles ZSSD through a two-stage pipeline that combines schema abstraction with graph-based neural inference. 
%% 0801
% The overall framework of CIRF is illustrated in Figure~\ref{fig:SEGKM}. 
Specifically, as shown in Figure \ref{fig:framework}, CIRF consists of: (1) an USI module, which leverages large-scale unlabeled data and LLMs to automatically induce a library of abstract, multi-relational reasoning schemas; and (2) a SEGKM, which parses each input argument into a FOL graph, aligns it with the induced schemas, and predicts stance via learnable graph kernel matching. 
% This design enables CIRF to generalize abstract reasoning patterns to novel and diverse targets while maintaining interpretability and robustness.

% \subsection{Task Definition}
\textbf{Task Definition.}
Let $X={(x_i,q_i)}_{i=1}^N$ represent the labeled data collection, where $x$ refers to the input text and $q$ corresponds to the source target. 
$N$ denotes the total number of instances in $X$. 
Each sentence-target pair $(x,q) \in X$ is assigned a stance label $y$ (e.g., \textit{favor}, \textit{against}, or \textit{none}). 
ZSSD aims to train a model from source known targets and predict the stance polarity towards unseen targets.

\subsection{USI: Unsupervised Schema Induction}

To bridge the gap between specific instance-level reasoning and general, transferable inference patterns, we design an unsupervised schema induction pipeline that extracts abstract cognitive schemas from LLM-generated logical reasoning.
Unlike prior approaches that directly cluster raw logical forms, our method involves a multi-stage reasoning analysis to ensure the resulting schemas are both meaningful and generalizable. 
Our pipeline consists of four main stages as follows. %, and all prompt templates and examples are provided in the Technical Appendix\footnote{https://github.com/Pluto132/CIRF/blob/main/Appendix.pdf}.

\textbf{FOL Reasoning Generation.}
For each sentence-target pair, we first prompt an LLM to generate a reasoning chain in the form of FOL. 

\textbf{FOL Interpretation and Abstraction.}
We further prompt the LLM to analyze the internal structure and inference logic of each generated FOL chain, and to produce alternative but logically equivalent FOL expressions that exhibit diverse syntactic or structural forms.
By collecting these logically similar yet structurally varied formulations, we are able to expose the underlying reasoning strategy beyond surface realization. 
Subsequently, we prompt the LLM to summarize the shared reasoning pattern among these variants into a generalized FOL template, which captures the essential logic in a reusable and domain-independent manner.
For example, consider the following case: 
$\forall x\ $, (is robot(x) $\rightarrow$ (helps humans(x) $\rightarrow$ (must be safe(x) $\land$ requires testing(x)))) $\rightarrow$ recommend regulation. 
After interpretation and abstraction, it will be generalized as:
$\forall x\ $, ((is target(x) $\land$ meets condition(x)) $\rightarrow$ entails consequence(x)) $\rightarrow$ policy action.

\textbf{Schema Clustering and Hierarchical Abstraction.}
We cluster the abstracted FOL templates into logical categories (schemas) by prompting the LLM to group them according to both semantic and inference pattern similarity. For each resulting schema, if a logic form does not fit any existing category, the LLM creates a new schema with a unique ID, name, and description. 
To capture the shared reasoning structure within each cluster, we prompt the LLM to synthesize a representative logic chain using both the cluster's description and its FOL members. For large clusters, we adopt a hierarchical strategy: FOLs are first grouped into fine-grained sub-clusters, each summarized into an intermediate logic chain, and these summaries are finally merged to form the schema-level template.

\textbf{Schema Graph Construction.}
Finally, we represent the induced schemas as nodes in a multi-relational graph, with edges indicating logical relations such as causality, contrast, or implication.
This multi-stage abstraction process distills diverse reasoning instances into a compact, interpretable, and transferable set of schema-level templates, enabling CIRF to robustly generalize reasoning patterns to previously unseen targets.

% \begin{figure}[t] % h 表示尝试将图片放在当前位置，顶，底，单独页面
%     \centering % 居中显示图片
%     \includegraphics[width=0.48\textwidth]{AnonymousSubmission/LaTeX/figures_0923/segkm.png}
%     \caption{The framework of SEGKM.}
%     \label{fig:SEGKM} % 设置图片标签，用于引用
% \end{figure}

\subsection{SEGKM: Schema-Enhanced Graph Kernel Model}

To leverage both instance-specific logical reasoning and abstract schema knowledge for zero-shot stance detection, we propose the SEGKM, as illustrated in Figure \ref{fig:framework}. 
SEGKM is designed to integrate transferable, concept-level cognitive schemas directly into the reasoning process of each test instance. These schemas are induced from LLM rationales via the USI module.
This enables the model to systematically align local reasoning structures with global, generalizable patterns, thus supporting interpretable and robust inference across unseen stance targets.
While standard GNNs primarily rely on local message passing, SEGKM explicitly matches input reasoning structures to abstract schema templates, facilitating structure-level generalization and interpretability.
Specifically, schema-driven reasoning in SEGKM proceeds as follows:

\textbf{FOL Graph Construction.}
Given an input sentence-target pair $(x, q)$, we first generate step-by-step reasoning rationales using LLM prompting (as in USI), and convert them into a FOL graph $G^f=(V^f, E^f)$, where nodes represent predicates and edges encode logical relations extracted from the input, such as implication, conjunction, or negation, corresponding to common FOL constructs.

\textbf{Schema Knowledge as Graph Filters.}
To inject abstract reasoning motifs into node-level feature extraction, we initialize the graph kernel filters using local subgraphs of the induced cognitive schemas. 
Specifically, for each schema graph $G^{(j)}$, we extract a set of subgraph filters ${H}^{(j)} = \{H_1^{(j)}, \ldots, H_{n}^{(j)}\}$, where each filter $H_i^{(j)}$ is a subgraph centered on a schema node, retaining its neighborhood structure and relational context.
Collecting filters from all schema graphs yields a filter pool $\mathcal{H} = \bigcup_j {H}^{(j)}$, where each group ${H}^{(j)}$ corresponds to a coherent, high-level reasoning schema.
Compared to using the full schema graph, these subgraph filters efficiently capture fine-grained relationships while reducing computation, and enable the kernel network to match localized reasoning structures from input graphs to reusable schema patterns.

\textbf{Kernel-Based Node Representation.}
To infuse schema-level reasoning patterns into the FOL-based input graph, we propose a kernel-based node representation method that integrates both structural and semantic similarity between the local subgraph around each node and pre-induced schema filters.
Each node and edge embedding is initialized via a pretrained BERT encoder~\cite{devlin-etal-2019-bert}, ensuring rich semantic information.

Let $G_v^f$ denote the $k$-hop local subgraph centered at node $v \in V^f$, which serves as the local reasoning structure for $v$.
To evaluate how well $G_v^f$ aligns with a schema filter $H_i^{(j)}$, we define a semantic-aware deep kernel function inspired by the $p$-step random walk kernel~\cite{feng2022kergnns}, enhanced with edge features.
Specifically, given the set of schema graphs $G^{(j)}$ and their subgraph filters $H_i^{(j)}$ constructed as described previously, 
inspired by~\citet{yin2024textgt}, we first encode edge semantics into node features via a relation-aware projection. For each node embedding $x$ and its associated edge feature $e$, we define:
\begin{equation}
x' = \text{ReLU}\left(x + \text{Proj}(e)\right), \quad e \in \mathbb{R}^{|E|}
\end{equation}
where $\text{Proj}(e)$ denotes a linear transformation of edge-type or relation embeddings. This mechanism ensures that both structural and relational patterns are captured during kernel evaluation. This operation is applied over all nodes in $G_v^f$ and $H_i^{(j)}$, yielding semantic-aware node representations $\mathbf{X}'_{G_v^f}$ and $\mathbf{X}'_{H_i^{(j)}}$.

Next, we compute a similarity matrix based on these refined node features:
\begin{equation}
S = \mathbf{X}'_{G_v^f} \cdot (\mathbf{X}'_{H_i^{(j)}})^\top
\end{equation}

Let $\mathbf{s} = \text{vec}(S) \in \mathbb{R}^{mn}$ denote the flattened similarity vector between all node pairs in $G_v^f$ and $H_i^{(j)}$. We then define the deep kernel response as:
\begin{equation}
\phi_{1,i}(v) = K_p(G_v^f, H_i^{(j)}) = \mathbf{s}^\top W A_{\times}^p \mathbf{s}
\end{equation}
Here, $A_{\times}^{p}$ denotes a learnable $p$-step transition matrix over the product graph of $G_v^f$ and $H_i^{(j)}$, which captures higher-order structural alignment, 
%% 0801
and $W$ is a learnable weight matrix that parametrizes the kernel response. 
This explicit product graph kernel formulation enables direct modeling of complex structural correspondences between input and schema.
In contrast, standard GNNs mainly rely on local message passing and often struggle to capture reusable, high-level reasoning motifs.
Note that while edge semantics are not directly injected into $A_{\times}^{p}$, they implicitly influence kernel computation through the edge-aware node features used in $S$.

Rather than selecting individual schema filters, we group filters by their originating schema graph and perform selection at the graph level. 
This grouping further reinforces the transferability and interpretability of schema-guided reasoning.

% Specifically, for each schema graph $G^{(j)}$, we aggregate the kernel responses of all its constituent filters $\mathcal{H}^{(j)} = \{H_1^{(j)}, \ldots, H_{n_j}^{(j)}\}$ with respect to the local subgraph $G_v^f$ to obtain an overall alignment score:
To compute the overall alignment between the input graph $G^f$ and a schema graph $G^{(j)}$, we aggregate the average kernel responses of all constituent filters $H_i^{(j)}$ across every node $v \in V^f$:
\begin{equation}
% \text{Score}(G^{(j)}) = \frac{1}{|\mathcal{H}^{(j)}|} \sum_{H_i^{(j)} \in \mathcal{H}^{(j)}} \phi_{1,i}(v)
S^{(j)} = \sum_{v \in V^f} \left( \frac{1}{|{H}^{(j)}|} \sum_{H_i^{(j)} \in {H}^{(j)}} \phi_{1,i}(v) \right)
\end{equation}
% Here, $\text{Score}(G^{(j)})$ represents the average alignment score over all filters $H_i^{(j)}$ of schema graph $G^{(j)}$, where each $\phi_{1,i}(v)$ measures the local kernel response between $G_v^f$ and filter $H_i^{(j)}$.
It ensures that the schema score reflects both local compatibility at each node and global coverage across the graph, capturing the degree to which the reasoning pattern embedded in $G^{(j)}$ manifests throughout the input structure.

% We then select the top-$g$ schema graphs based on these cumulative scores:
% We then select the top-$g$ schema graphs based on their overall alignment with the input graph $G^f$:
% \begin{equation}
% % % \mathcal{G}^*(v) = \arg\max_{\mathcal{G}' \subseteq \mathcal{G},\, |\mathcal{G}'| = g} \sum_{H^{(j)} \in \mathcal{G}'} \text{Score}(H^{(j)})
% \mathcal{G}^* = \arg\max_{\mathcal{G}' \subseteq \mathcal{G},\, |\mathcal{G}'| = g} \sum_{G^{(j)} \in \mathcal{G}'} \text{Score}(G^{(j)})
% \end{equation}
% where $\mathcal{G}$ denotes the full set of schema graphs; $\mathcal{G}'$ is a candidate subset of $\mathcal{G}$ with exactly $g$ schema graphs; $H^{(j)}$ refers to the $j$-th schema graph in $\mathcal{G}'$.
We then select the top-$g$ schema graphs with the highest scores $S^{(j)}$ as the candidate schema set $\mathcal{G}^*$, which will provide the filter set for node representation. 
The filters associated with these selected schema graphs are aggregated to form the candidate reasoning motifs used for downstream node representation learning.
% \begin{equation}
% \mathcal{G}^* = \arg\max_{\mathcal{G}' \subseteq \mathcal{G},\, |\mathcal{G}'| = g} \text{Score}(G^{(j)})
% \end{equation}
% where $\mathcal{G}$ denotes the full set of schema graphs; $\mathcal{G}'$ is a candidate subset of $\mathcal{G}$ with exactly $g$ schema graphs; $H^{(j)}$ refers to the $j$-th schema graph in $\mathcal{G}'$.

Finally, we construct the node representation $\phi_1(v)$ by concatenating the responses $\phi_{1,i}(v)$ for all filters $H_i^{(j)}$ belonging to the selected top-$g$ schema graphs:
\begin{equation}
% \phi_1(v) = \text{Concat}(\phi_{1,i}(v) \mid H_i^{(j)} \in \bigcup_{H^{(j)} \in \mathcal{G}^*(v)} {H}^{(j)})
\phi_1(v) = \text{Concat}\left( \phi_{1,i}(v) \,\middle|\, H_i^{(j)} \in \bigcup_{G^{(j)} \in \mathcal{G}^*} {H}^{(j)} \right)
\end{equation}

This yields a schema-aware node representation that reflects the most semantically and structurally relevant reasoning motifs. 
By explicitly aligning local reasoning patterns with high-level schemas through the graph kernel, our method supports interpretable and transferable inference, which is particularly advantageous for ZSSD.

% \textbf{Multi-layer Model.}
% \textbf{Multi-hop Reasoning via Schema-Kernel Layers}
% \textbf{Graph Representation via Stacked Schema-Kernel Layers.}
\textbf{Hierarchical Kernel-Based Graph Representation.}
Real-world stance reasoning often requires multi-hop logic and the flexible composition of multiple schema patterns. 
To support this, we stack multiple layers of schema-driven kernel feature extraction. 
At each layer $l$, node representations are recursively updated by matching their expanded local subgraphs to the schema filters, allowing deeper layers to aggregate schema knowledge from increasingly broader reasoning contexts.
This design enables SEGKM to capture both fine-grained local logic and higher-order, composite reasoning motifs.
% —crucial for robust generalization in zero-shot stance detection.
The final graph representation is constructed by concatenating the aggregated node features from all layers:
\begin{equation}
    \Phi(G^f) = \text{Concat}\left(\sum_{v \in G^f} \phi_l(v) \mid l = 0, 1, ..., L\right)
\end{equation}

The final graph representation is used for stance prediction, and the model is trained end-to-end using cross-entropy loss.

% \subsection{Prediction and Loss}

% The final graph representation $\Phi(G)$ is fed into a fully connected layer with softmax to produce the stance prediction:
% \begin{equation}
% \hat{y} = \text{softmax}(W_{\text{o}} \cdot \Phi(G) + b_{\text{o}})
% \end{equation}
% where $W_{\text{o}}$ and $b_{\text{o}}$ are trainable parameters. We train the model using cross-entropy loss.

\section{Experimental Setups}
% \textbf{Experimental Data.}
% We evaluate our approach on three widely used stance detection benchmarks: SEM16, VAST, and COVID-19-Stance.

% {SEM16} (SemEval-2016 Task 6;~\cite{mohammad2016semeval}) is a benchmark of tweets annotated for stance towards six predefined targets. Following prior work~\cite{li2023stance,lan2024stance}, we focus on three commonly used targets: \textit{Hillary Clinton} (HC), \textit{Feminist Movement} (FM), and \textit{Legalization of Abortion} (LA).

% {VAST}~\cite{allaway2020zero} comprises texts from the New York Times ``Room for Debate'' section, covering a broad spectrum of topics. The dataset contains 4,003 training, 383 development, and 600 test examples, with topics ranging from education and politics to public health. Topic phrases are automatically extracted and subsequently refined by human annotators to ensure quality and diversity.

% {COVID-19} (COVID-19-Stance;~\cite{glandt2021stance}) is constructed to assess public stances toward COVID-19-related policies. It includes four targets: \textit{Wearing a Face Mask} (WA), \textit{Keeping Schools Closed} (SC), \textit{Anthony S. Fauci, M.D.} (AF), and \textit{Stay at Home Orders} (SH).

\textbf{Experimental Data.}
We evaluate CIRF on three widely used stance detection benchmarks:
\textbf{SEM16} (SemEval-2016 Task 6;~\cite{mohammad2016semeval}) consists of tweets annotated for stance towards three targets: \textit{Hillary Clinton} (HC), \textit{Feminist Movement} (FM), and \textit{Legalization of Abortion} (LA).
\textbf{VAST}~\cite{allaway2020zero} contains New York Times opinion texts covering diverse topics, with 4,003 training, 383 development, and 600 test examples. Topic phrases are refined by human annotators to ensure quality.
\textbf{COVID-19} (COVID-19-Stance;~\cite{glandt2021stance}) includes data on four COVID-19-related targets: \textit{Wearing a Face Mask} (WA), \textit{Keeping Schools Closed} (SC), \textit{Anthony S. Fauci, M.D.} (AF), and \textit{Stay at Home Orders} (SH).
All datasets follow standard splits and label settings from previous work.
% For detailed dataset statistics, see Technical Appendix B.

\textbf{Evaluation Metrics.}
For the SEM16 and COVID-19 datasets, which include three classes (``FAVOR'', ``AGAINST'', and ``NONE''), we follow~\citet{liang2022zero} and report macro-F1 ($F_{avg}$) computed over the ``FAVOR'' and ``AGAINST'' categories. 
For the VAST dataset, which comprises three categories (``Pro'', ``Con'', and ``Neutral''), we follow~\citet{li2023tts} and report the macro-F1 across all classes, along with individual macro-F1 scores for the ``Pro'' and ``Con'' categories.

\begin{table*}[ht]
    \centering
    \small
    % \resizebox{\linewidth}{!}{
    \begin{tabular}
    % {c|c|ccc|ccc|ccc|cccc}
    {cc ccc ccc ccc cccc}
    \toprule
   \multirow{3}{*}{}  & \multirow{3}{*}{Model} & \multicolumn{3}{c}{SEM16} & \multicolumn{3}{c}{VAST (100\%)} & \multicolumn{3}{c}{VAST (10\%)} & \multicolumn{4}{c}{COVID-19} \\
\cmidrule(lr){3-5} \cmidrule(lr){6-8} \cmidrule(lr){9-11} \cmidrule(lr){12-15}
& & HC & FM & LA & Pro & Con & All & Pro & Con & All & AF & SC & SH & WA \\ \midrule
    % \multirow{1}{*}
    {Glove}
    % & BiLSTM     & 31.6  & 40.3  & 33.6  & 43.7  & 43.6  & 37.5  & 39.2  & 24.6  & 32.9  & 26.7  & 33.9  & 19.3  & 30.1  \\
    & CrossNet   & 38.3  & 41.7  & 38.5  & 46.2  & 43.4  & 43.4  & 37.3  & 32.9  & 36.2  & 41.3  & 40.0  & 40.4  & 38.2  \\
    % & SEKT       & 50.1  & 44.2  & 44.6  & 50.4  & 44.2  & 41.8  & -     & -     & -     & - & - & - & - \\
    \cdashline{1-15}[2pt/3pt]
    \multirow{3}{*}
    {Bert}
    & JointCL    & 54.4  & 54.0  & 50.0  & 64.9  & 63.2  & 71.2  & 53.8  & 57.1  & 65.5  & 57.6$^\dagger$ & 49.3$^\dagger$ & 43.5$^\dagger$ & 63.1$^\dagger$ \\
    & TarBK      & 55.1  & 53.8  & 48.7  & 65.7  & 63.9  & 73.6  & -     & -     & -     & - & - & - & - \\
    & PT-HCL     & 54.5  & 54.6  & 50.9  & 61.7  & 63.5  & 71.6  & -     & -     & -     & - & - & - & - \\
    % & NPS4SD     & 60.1  & 56.7  & 51.0  & 69.2  & 67.6  & 74.4  & 68.7  & 64.3  & 72.0  & - & - & - & - \\
    \cdashline{1-15}[2pt/3pt]
    \multirow{9}{*}{GPT-3.5}
    & GPT-3.5     & 78.9  & 68.3  & 62.3  & 63.8  & 56.8  & 65.1  & 63.8  & 56.8  & 65.1  & 69.2$^\dagger$ & 43.5$^\dagger$ & 66.5$^\dagger$ & 57.8$^\dagger$ \\
    & COLA       & \underline{81.7}  & 63.4  & \underline{71.0}  & -     & -     & 73.0  & -     & -     & 73.0  & 65.7$^\dagger$ & 46.6$^\dagger$ & 53.5$^\dagger$ & 73.9$^\dagger$ \\
    % 77.64 72.47 77.24 59.20
    % & KASD-BERT       & 64.8  & 57.1  & 51.6  & 62.5  & 63.8  & 76.9  & -     & -     & -     & - & - & - & - \\
        & KASD       & 80.3  & 70.4  & 62.7 & -  & -  & 67.0  & -     & -     & -     & - & - & - & - \\
    % & MultiPLN   & 77.6  & \textit{\color{red}{74.8}}  & 68.8  & 69.3  & 67.9  & 74.8  & -     & -     & -     & - & - & - & - \\
    % & EDDA       & 77.4  & 69.7  & 62.7  & 66.9  & 68.2  & 75.1  & 62.6  & 66.9  & 72.7  & - & - & - & - \\
    & KAI        & 76.4  & \underline{73.7}  & 69.4  & 66.7  & 73.0  & 76.3  & 63.5  & 74.0  & 75.2  & - & - & - & - \\
    % zhang2025zero
    & LCDA       & 79.8  & 70.0  & 69.4  & \underline{73.8}  & \underline{75.9}  & \underline{80.3}  & \underline{72.1}  & \underline{77.0}  & \underline{80.1}  & - & - & - & - \\
    % zhang2023logically
    % & LC-CoT & 82.9 & 70.4 & 63.2 & -  & -  & 72.5  & -  & -  & - & - & - & - & - \\ 
    % dai2025large
    & FOLAR$^\dagger$ & \textbf{81.9} & 71.2 & 69.9 & 71.2 & 75.8 & 77.2 & 70.1 & 75.5 & 76.7 & 69.5 & 67.2 & \underline{65.4} & 73.1 \\
    % zhang2025logic
    & LogiMDF$^\dagger$       & 75.1  & 67.9  & 68.0  & -  & -  & 76.7  & -  & -  & 76.6  & \underline{70.4} & \underline{68.8} & 64.9 & \underline{75.4} \\ 
     % \cdashline{2-15}[2pt/3pt]
     
   % & 
   \cdashline{2-15}[2pt/3pt]
   &  {\textbf{CIRF}}$^\ddag$ & {80.1} & \textbf{74.7} & \textbf{73.9} & \textbf{74.1} & \textbf{78.5} & \textbf{80.9} & \textbf{73.4} & \textbf{78.4} & \textbf{80.7} & \textbf{74.1} & \textbf{70.3} & \textbf{68.8} & \textbf{81.0} \\
   % & \textbf{CIRF} & 80.1 & \textbf{74.7}$^\ddag$ & \textbf{73.9}$^\ddag$ & \textbf{74.1}$^\ddag$ & \textbf{78.5}$^\ddag$ & \textbf{80.9}$^\ddag$ & \textbf{73.4}$^\ddag$ & \textbf{78.4}$^\ddag$ & \textbf{80.7}$^\ddag$ & \textbf{74.1}$^\ddag$ & \textbf{70.3}$^\ddag$ & \textbf{68.8}$^\ddag$ & \textbf{81.0}$^\ddag$ \\

 % \cmidrule{1-15}
 \hline
   
      \multicolumn{2}{r}{DeepSeek-v3} & 80.7 & 78.3 & 77.7 &  75.0 & 74.3  & 75.9 &  75.0 & 74.3  & 75.9  & 76.3 & 74.3 & 77.1 & 87.6  \\
       \multicolumn{2}{r}{GPT-4o} & 81.6 & 79.8 & 77.6 & 77.6  & 78.7  & 80.0 & 77.6  & 78.7  & 80.0  & 82.7 & 78.8 & 76.9 & 89.2 \\
    % \rowcolor[gray]{0.92}
    % \multicolumn{2}{r}{DeepSeek-v3} & 
    % 80.7 & 78.3 & 77.7 & 75.0 & 
    % 74.3 & 75.9 & 75.0 & 74.3 & 
    % 75.9 & 76.3 & 74.3 & 77.1 & 
    % 87.6 \\
    
    % \rowcolor[gray]{0.92}
    % \multicolumn{2}{r}{GPT-4o} & 
    % 81.6 & 79.8 & 77.6 & 77.6 & 
    % 78.7 & 80.0 & 77.6 & 78.7 & 
    % 80.0 & 82.7 & 78.8 & 76.9 & 
    % 89.2 \\
     \cdashline{1-15}[2pt/3pt]
    % \cdashline{2-15}[2pt/3pt]
    % \cmidrule{1-15}
    % & CIRF (Ours)       & 79.2     & \textbf{74.3}$^\ddag$     & \textbf{72.5}$^\ddag$   &  \textbf{74.1}  & \textbf{78.2}  & \textbf{80.8}$^\ddag$  & \textbf{73.1}  & \textbf{77.8}  & \textbf{80.4}$^\ddag$  & \textbf{73.9}$^\ddag$ & \textbf{70.0}$^\ddag$ & \textbf{68.0}$^\ddag$  & \textbf{80.8}$^\ddag$ \\
    % % \cmidrule{2-15}
    % & CIRF (+ se)      & 79.4  & 74.3 & 71.8  & 72.2  & 77.3  & 79.9  & 73.2  & 78.0  & 80.4  & 73.9 & 70.3 & 68.0 & 80.9 \\
    % \multirow{2}{*}{CIRF}
    % & GPT-3.5 & 80.0 & \textbf{74.5} & \textbf{73.1} & \textbf{74.1} & \textbf{78.5} & \textbf{80.9} & \textbf{73.2} & \textbf{78.0} & \textbf{80.4} & \textbf{74.0} & \textbf{70.3} & \textbf{68.5} & \textbf{80.9} \\
    % & CIRF (DeepSeek-v3) 
    % \rowcolor[gray]{0.92}
    % \multicolumn{2}{>{}r}{\textbf{CIRF(DeepSeek-v3)}} & 
    
    % \multicolumn{2}{r}{\textbf{CIRF(DeepSeek-v3)}} & 
    % {83.8}  & {79.8}  & {78.2}  & {76.4}  & {77.1}  & {80.3}  & {76.2}  & {76.8}  & {80.1}  & {78.5}  & {78.8}  & {78.3}  & {89.1} \\
    % % \rowcolor[gray]{0.92}
    % % \multicolumn{2}{>{}r}{\textbf{CIRF(GPT-4o)}}   
    % \multicolumn{2}{r}{\textbf{CIRF(DeepSeek-v3)}}
    % & {83.2}  & {80.4}  & {78.2}  & {78.8}  & {80.1}  & {82.8}  & {78.6}  & {79.6}  & {82.5}  & {84.9}  & {80.5}  & {78.6} & {89.4} \\

    \multicolumn{2}{r}{CIRF (DeepSeek-v3)} & 
    83.8  & 79.8  & 78.2  & 76.4  & 77.1  & 80.3  & 76.2  & 76.8  & 80.1  & 78.5  & 78.8  & 78.3  & 89.1 \\
    
    \multicolumn{2}{r}{CIRF (GPT-4o)} & 
    83.2  & 80.4  & 78.2  & 78.8  & 80.1  & 82.8  & 78.6  & 79.6  & 82.5  & 84.9  & 80.5  & 78.6 & 89.4 \\

    \bottomrule
    \end{tabular}
    % }
\caption{Comparison of different models on the ZSSD task. The best scores are highlighted in bold, and the second-best scores are underlined. 
$^\ddag$ indicates that CIRF outperforms FOLAR, LCDA, and LogiMDF in macro-F1 across targets (paired \textit{t}-test, $p < 0.05$).
% $^\ddag$ indicates statistically significant improvements of our CIRF over FOLAR based on paired \textit{t}-tests with p-value $<$ 0.05. 
$^\dagger$ denotes results reproduced using the official open-source code. The lower part of the table represents CIRF evaluated with more powerful LLM backbones to examine scalability.} 
    \label{zeroshot}
\end{table*}

\textbf{Implementation Details.}
We use 
GPT-3.5 % \footnote{\url{https://platform.openai.com/docs/models/gpt-3-5-turbo}} 
as the default LLM for FOL elicitation and summary generation, following prior baselines.
To evaluate robustness, we also test with 
GPT-4o % \footnote{\url{https://platform.openai.com/docs/models/gpt-4o}}
and DeepSeek-v3 %\footnote{\url{https://api-docs.deepseek.com/}}
. Dataset splits follow~\citet{li2023stance,lan2024stance}: for SEM16 and COVID-19, one target is held out for testing; for VAST, we use the official zero-shot setup. 
% USI uses temperature 0 for LLM queries.
USI uses temperature 0 for LLM queries. 
% (see derived schemas in Appendix E).
% For the schemas derived in the experiments, 
% For the schemas derived via USI in the experiments, 
% see Technical Appendix E.
% For the schemas derived via USI in the experiments; see Technical Appendix E.
SEGKM adopts a two-layer graph kernel, and applies a fully connected ReLU layer. 
Training (on a 40GB A100 GPU) uses AdamW (batch size 32, learning rate $5\!\times\!10^{-4}$), early stopping (patience 10), up to 20 epochs, and selects the checkpoint with the lowest validation loss (validation every 0.2 epoch). We report the average result over three independent runs.

\textbf{Baseline Methods.}
To evaluate the performance of existing stance detection models on our dataset, we employed the following models:
(1) Non-LLM methods:
% {BiLSTM}~\cite{augenstein2016stance} and 
{CrossNet}~\cite{ijcai2017p557},
% (2) Fine-tuned model baselines:
 {JointCL}~\cite{liang2022jointcl}, 
 {TarBK}~\cite{zhu2022enhancing}, and 
{PT-HCL}~\cite{liang2022zero}.
% and {NPS4SD}~\cite{zhang2023twitter}.
(2) Zero-shot prompting LLMs: 
{GPT-3.5}~\cite{zhang2023investigating}, 
{COLA}~\cite{lan2024stance}, and
{KASD}~\cite{li2023stance}.
(3) LLM-enhanced fine-tuned models:
% {MultiPLN}~\cite{ding2024cross}, 
{KAI}~\cite{zhang2024knowledge},
{LCDA}~\cite{zhang2025zero},
{FOLAR}~\cite{dai2025large},
and {LogiMDF}~\cite{zhang2025logic}.

\section{Experimental Results}

% \subsection{Main ZSSD Experimental Results}
\paragraph{Main ZSSD Experimental Results.}
The main ZSSD results are summarized in Table~\ref{zeroshot}. CIRF consistently outperforms all baseline models across all three datasets, demonstrating the effectiveness of our approach in challenging zero-shot scenarios.
For example, CIRF achieves an average F1 score of 76.2 on SEM16 and 80.9 on VAST, representing a 1.9-point improvement over FOLAR and a 0.6-point gain over LCDA, respectively, and surpassing LogiMDF, the most competitive method on the COVID-19 dataset, by 3.7 points. Statistical significance tests (p $<$ 0.05) confirm that CIRF's improvements over FOLAR and LogiMDF are significant.

A closer examination reveals several trends:
First, traditional non-LLM methods perform substantially worse on ZSSD, while LLM-based models yield notable gains, emphasizing the crucial role of LLMs' reasoning capabilities.
Furthermore, fine-tuned LLM-based models (KAI, LogiMDF, FOLAR, and CIRF) generally surpass direct prompting strategies (GPT-3.5, COLA, and KASD), suggesting that combining annotated data with the knowledge encoded in LLMs is more effective.

Breaking down by dataset, CIRF’s advantage is most pronounced on VAST, which features greater topic diversity and fine-grained targets, showcasing its robustness in real-world scenarios. On SEM16 and COVID-19, where the targets are more constrained, all LLM-based models perform more closely, but CIRF still consistently ranks first.

We also compare different knowledge representations.
Models using FOL knowledge elicited from LLMs (CIRF and FOLAR) generally outperform those using natural language (KAI), indicating that FOL provides a more compact and effective abstraction for reasoning transfer. Notably, CIRF further outperforms KAI, the strongest prior LLM-based method, highlighting the added value of our cognitive framework enhancement.

These results collectively demonstrate that CIRF not only sets a new state-of-the-art for ZSSD but also provides insights into the importance of structured LLM-elicited knowledge and cognitive schema design for robust zero-shot stance reasoning.
Overall, this validates our hypothesis that explicit schema-guided abstraction enables better generalization to unseen targets and diverse domains.

To further investigate the scalability of CIRF, we evaluate it with stronger LLM backbones, including DeepSeek-v3 and GPT-4o. CIRF’s performance rises notably with these advanced models: for example, switching from GPT-3.5 to GPT-4o increases the average F1 on VAST  from 80.9 to 82.8, and on WA from 81.0 to 89.4. Similar gains are observed with DeepSeek-v3, confirming that CIRF can effectively capitalize on the enhanced reasoning capabilities of newer LLMs. These results highlight CIRF’s scalability and suggest that combining schema-guided abstraction with continually improving LLMs is a promising direction for future zero-shot stance detection research.

\begin{figure}[t] % h 表示尝试将图片放在当前位置，顶，底，单独页面
    \centering % 居中显示图片
    \includegraphics[width=\linewidth]{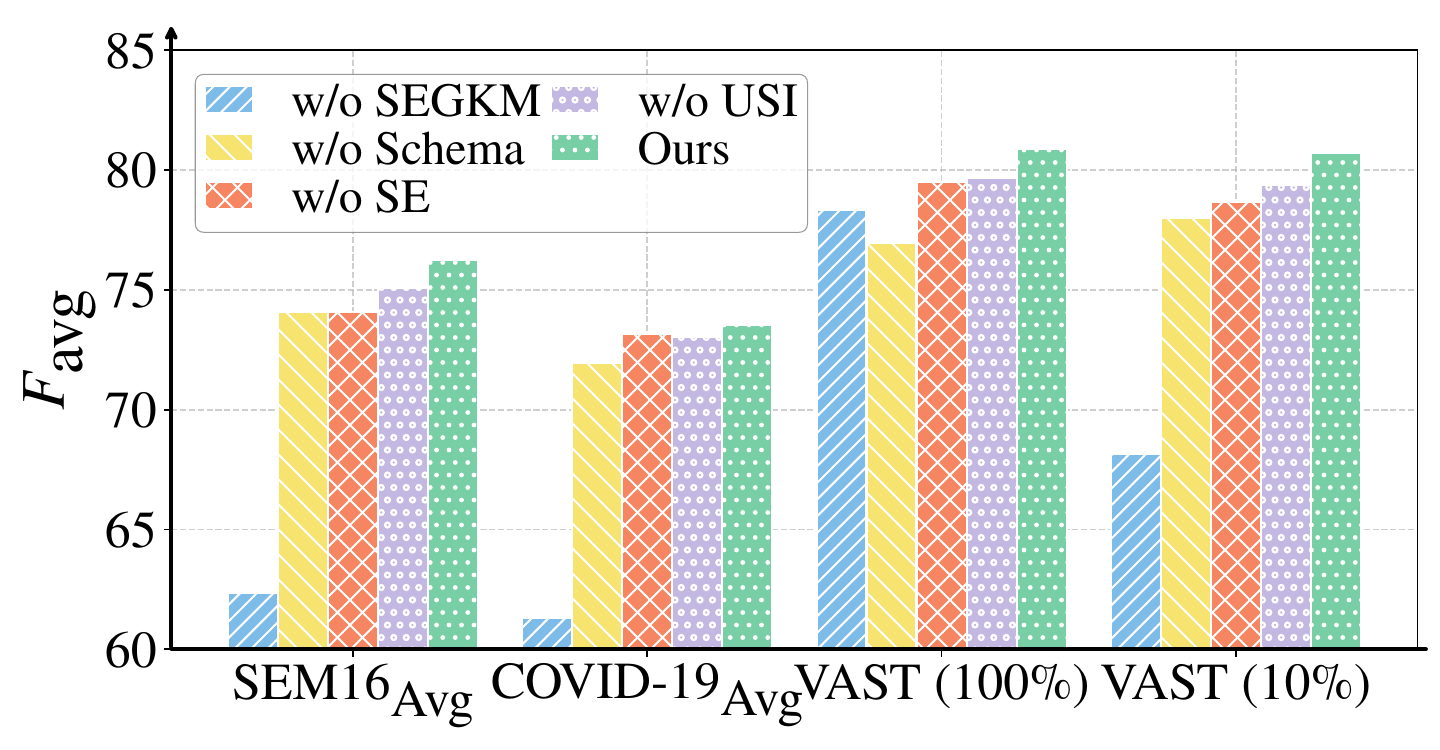}
    \caption{Ablation study results. Here, $_{Avg}$ denotes the average performance.}
    % \label{cszssd} % 设置图片标签，用于引用
    \label{fig:merge_ablation}
\end{figure}

% \subsection{Ablation Study}
\paragraph{Ablation Study.}
Figure~\ref{fig:merge_ablation} presents ablation results on SEM16, COVID-19, and VAST, quantifying the contribution of each key component in CIRF. Removing the cognitive schema (\textit{w/o Schema}) or the SEGKM (\textit{w/o SEGKM}) leads to the most substantial performance drops across all benchmarks, underscoring the central role of explicit schema-guided reasoning and relational logic encoding in robust zero-shot transfer. We also observe that eliminating edge semantics (\textit{w/o SE}) or replacing LLM-based schema induction with simple clustering (\textit{w/o USI}) consistently degrades results, though to a lesser extent, highlighting the importance of both rich relational structure and semantically grounded schema construction. Notably, the performance gaps become even wider under low-resource settings such as VAST (10\%), emphasizing that each component is critical for generalization in challenging domains. Overall, these findings confirm that the synergy of schema abstraction, semantic relations, and LLM-driven structure underpins CIRF’s effectiveness in zero-shot stance detection.
% For target-wise ablation results on SEM16 and COVID-19 datasets, please refer to Technical Appendix C.

\begin{figure}[t] % h 表示尝试将图片放在当前位置，顶，底，单独页面
    \centering % 居中显示图片
    \includegraphics[width=\linewidth]{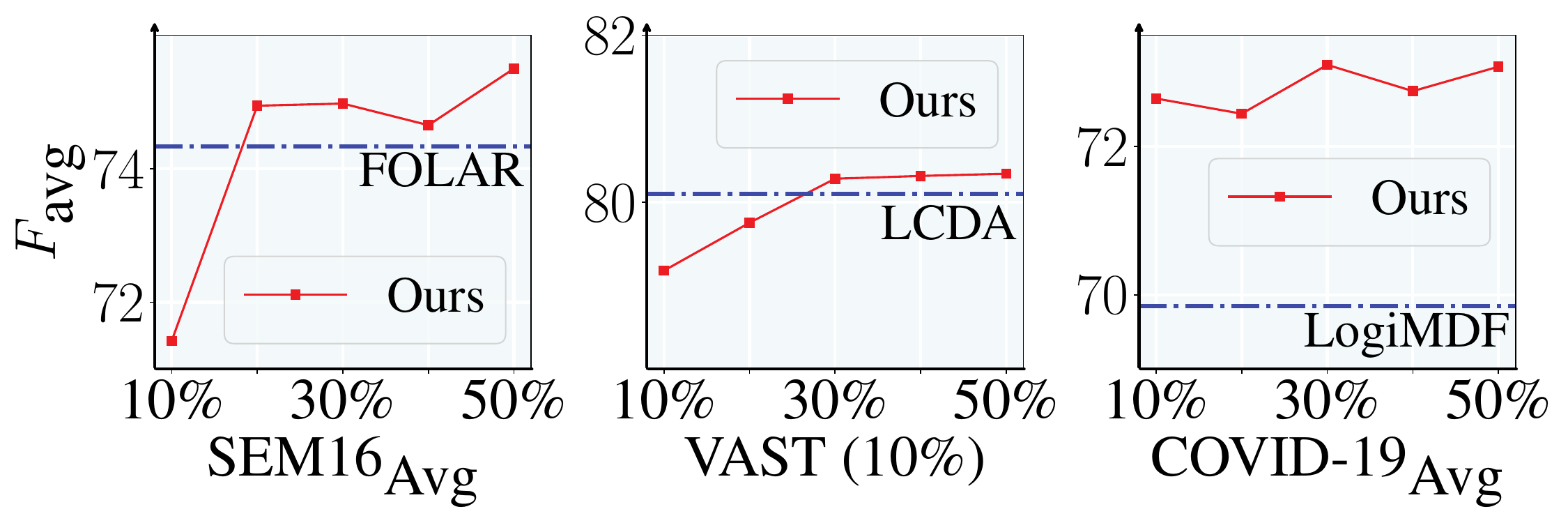}
    \caption{Performance comparison across different training scales. Dashed lines denote the strongest baseline method for each dataset.
    % on the SEM16, VAST (10\%), and COVID19 datasets. On the SEM16 and VAST (10\%) datasets, COLA and FOLAR serve as the strongest baselines for the ZSPM and LEM methods, respectively. For the COVID19 dataset, LogiMDF demonstrates the best performance among competing methods.
    }
    % \label{cszssd} % 设置图片标签，用于引用
    \label{fig:scale}
\end{figure}

\paragraph{Low-Resource Performance.}
To assess data efficiency under limited supervision, we evaluate CIRF on VAST (10\%), SEM16, and COVID-19 with varying proportions of labeled data. As shown in Figure~\ref{fig:scale}, CIRF consistently and substantially outperforms strong LLM-based baselines (FOLAR, LCDA, LogiMDF) across all training scales.
On COVID-19, CIRF outperforms LogiMDF by 2.8 points with only 10\% of supervision. Similarly, on SEM16, CIRF exceeds FOLAR by 0.6 points with just 20\% of labeled data. On VAST (10\%), CIRF steadily improves with more supervision and begins to outperform LCDA at 30\%. The performance gap against most baselines remains stable or widens as supervision increases.
% }
These results demonstrate that the abstract cognitive schema, constructed from unlabeled data in other domains, enables CIRF to achieve strong cross-domain generalization with minimal supervision. Our findings highlight the power of explicit, domain-agnostic schema abstraction for robust and data-efficient stance detection in low-resource settings.

\begin{figure}[h] % h 表示尝试将图片放在当前位置，顶，底，单独页面
    \centering % 居中显示图片
    \includegraphics[width=0.47\textwidth]{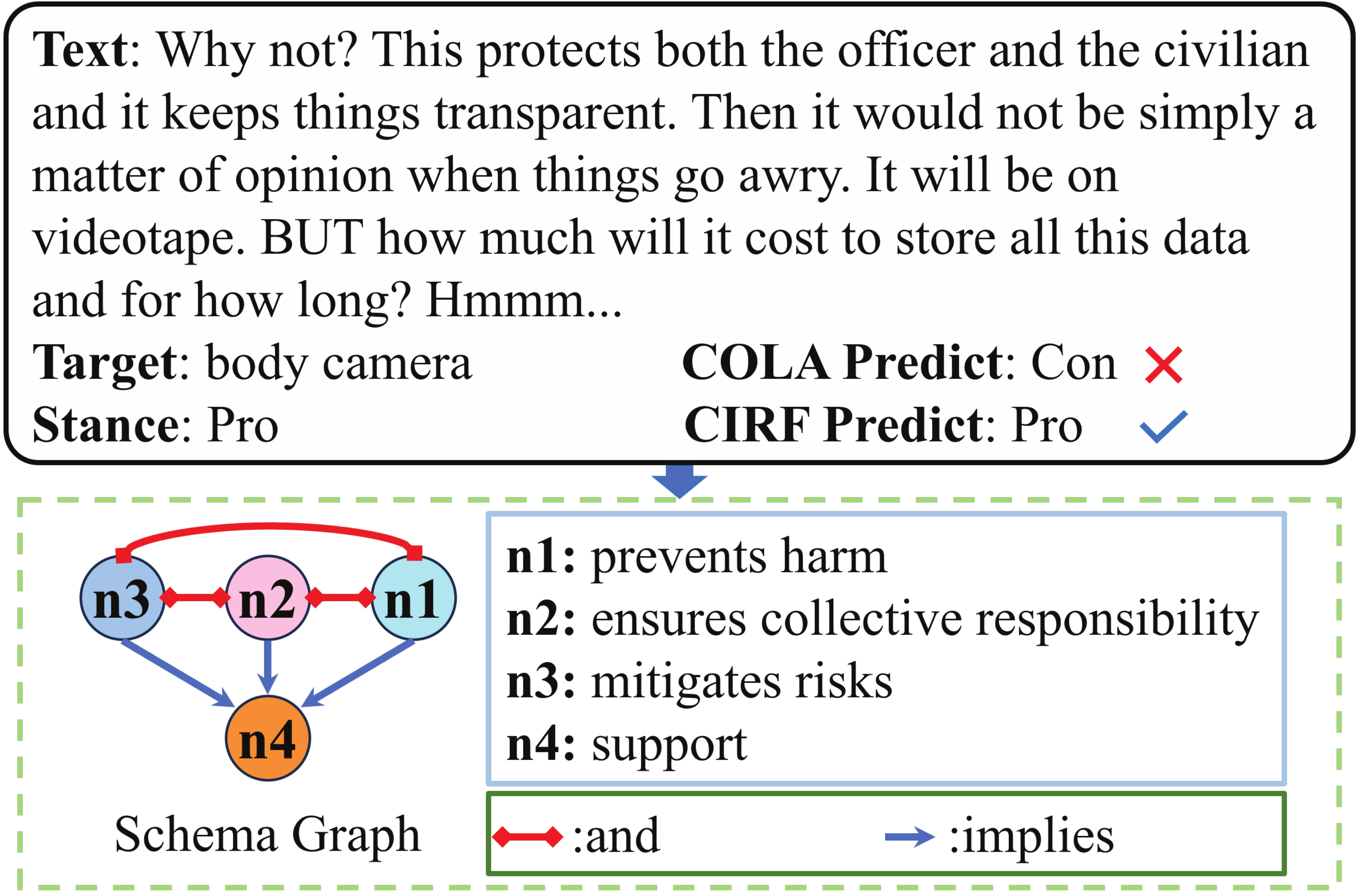}
    \caption{Case Study.}
    \label{fig:case_study_pro_bc} % 设置图片标签，用于引用
\end{figure}

\paragraph{Case Study.}
We present an example illustrating how schema-guided graph reasoning enables correct stance prediction under mixed sentiment (Figure~\ref{fig:case_study_pro_bc}). While the text overall supports body cameras, it contains a hypothetical cost concern introduced by an adversative “BUT.” Baseline models may be misled by this structure and overemphasize the cost issue, leading to misclassification. In contrast, our schema graph aligns key arguments with salient, domain-agnostic nodes such as \textit{prevents harm}, \textit{ensures collective responsibility}, and \textit{mitigates risks}. This structured reasoning allows CIRF to robustly capture the author’s supportive stance, even in nuanced or hedged cases.
% \textcolor{blue}{
% Additional case studies demonstrating similar graph-aligned reasoning appear in Technical Appendix D.
% }

\begin{figure}[t] % h 表示尝试将图片放在当前位置，顶，底，单独页面
    \centering % 居中显示图片
    \includegraphics[width=0.47\textwidth]{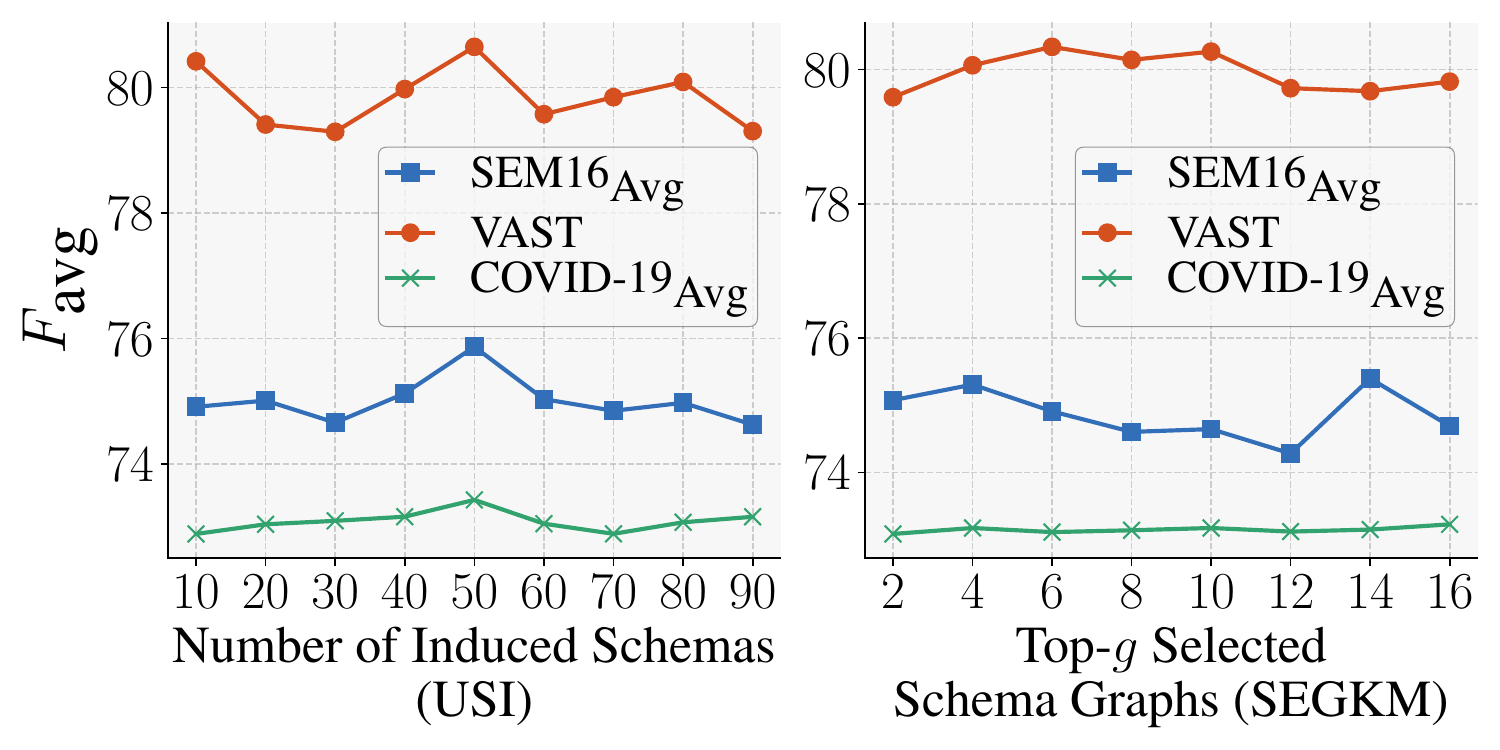}
    \caption{Impact of schema number (left) and selected schema graph number (right).
    % Impact of Schema Number in USI and Selected Schema Graph Number in SEGKM.
    }
    \label{schema_num} % 设置图片标签，用于引用
\end{figure}

\paragraph{Effect of the Number of Induced Schemas.}
We investigate the effect of varying the number of schemas on CIRF's zero-shot stance detection performance. As illustrated in Figure~\ref{schema_num}, the model's performance remains highly stable across a wide range of schema counts, with only minor fluctuations (generally less than 1 point) observed for each dataset. This robustness indicates that CIRF is largely insensitive to the schema number: a moderate schema set is sufficient to capture key reasoning patterns, and further increasing the number yields little additional benefit or risk of instability. These results support our hypothesis that stance reasoning can be well-abstracted by a structured, logic-inspired schema set, and demonstrate that CIRF is practical and easy to tune in real-world scenarios.

\paragraph{Effect of Top-$g$ Schema Graph Selection.}
We analyze how varying the number of schema graphs dynamically selected during inference affects CIRF's zero-shot stance detection performance. As shown in Figure~\ref{schema_num}, performance across all datasets remains highly stable as the number of selected schema graphs increases from 2 to 16, with only minor fluctuations observed for each dataset. The absence of any clear trend or performance drop demonstrates that CIRF is robust to this hyperparameter, and that a modest number of selected schema graphs is sufficient to capture the essential reasoning patterns for each instance. Increasing the number beyond this range offers little additional benefit. These findings support our design hypothesis that stance reasoning can be abstracted by a compact set of logic-based schema graphs, and further simplify model tuning and deployment, confirming that CIRF is practical and easy to adapt to diverse domains and resource settings.

\section{Conclusion}
In this work, we introduce the CIRF, a schema-driven approach for zero-shot stance detection that bridges linguistic input and abstract reasoning through automatic schema induction and schema-guided inference. By leveraging unsupervised abstraction of first-order logic patterns and schema-enhanced graph kernel alignment, CIRF enables robust generalization and interpretability across diverse and previously unseen targets. 
Extensive experiments on multiple benchmarks demonstrate that CIRF not only achieves new state-of-the-art performance, but also maintains competitive results with minimal labeled data, highlighting its practical value in low-resource settings.
Despite these advances, our framework still faces challenges, such as scaling schema induction for extremely large or noisy corpora, and integrating richer world knowledge or multimodal signals. 
Future work includes exploring more efficient schema induction algorithms, adaptive schema selection strategies, and extending CIRF to broader reasoning tasks in natural language understanding. 
Looking ahead, we will refine cognitive schema generation and extend CIRF to broader linguistic and cultural contexts to further improve its generalizability and real-world effectiveness.

\section*{Acknowledgements}
This research is supported by the National Natural Science Foundation of China No. 62306184, Natural Science Foundation of Top Talent of SZTU (grant no. GDRC202320), and Shenzhen Science and Technology Program (No. JCYJ20240813113218025).

\bibliography{aaai2026}

\end{document}